\pdfoutput=1
\documentclass{article}
\usepackage[dvipdfmx]{graphicx}
\usepackage{arxiv}

\usepackage[utf8]{inputenc} 
\usepackage[T1]{fontenc} 
\usepackage{hyperref}
\usepackage{url} 
\usepackage{amsmath,amsfonts} 
\usepackage{nicefrac}  
\usepackage{cleveref}
\usepackage{doi}

\usepackage{colortbl}
\usepackage{listings}

\usepackage{parskip}

\usepackage{booktabs}
\usepackage{threeparttable}
\usepackage{tabularx}
\usepackage{multirow}

\usepackage{seqsplit}
\newcommand{\ccol}[2]{ \multicolumn{#1}{c}{#2}}
\newcommand{\graycell}{\cellcolor[gray]{0.85}}

\usepackage{xparse}
\NewDocumentCommand{\citep}{o o m}{%
  \IfValueT{#1}{(#1) }\cite{#3}\IfValueT{#2}{ #2}%
}

\newcommand{\myPaperTitle}{Reference Points in LLM Sentiment Analysis: \\The Role of Structured Context}
\newcommand{\myPaperShortTitle}{Reference Points in LLM Sentiment Analysis: The Role of Structured Context}
\title{\myPaperTitle}
\date{}

\newif\ifuniqueAffiliation

\usepackage{authblk}

\newbox{\orcid}\sbox{\orcid}{\includegraphics[scale=0.06]{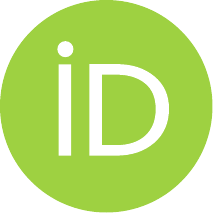}} 
\author[1,2]{%
	\href{https://orcid.org/0000-0002-4618-6272}{\usebox{\orcid}\hspace{1mm}
	Junichiro Niimi\thanks{\texttt{jniimi@meijo-u.ac.jp}}
	}}
\affil[1]{Meijo University}
\affil[2]{RIKEN AIP}

\renewcommand{\shorttitle}{\myPaperShortTitle}

\hypersetup{
pdftitle={Reference Points in LLM Sentiment Analysis: The Role of Structured Context},
pdfsubject={unsubjected},
pdfauthor={Junichiro ~Niimi},
pdfkeywords={Marketing, Natural Language Processing, Sentiment Analysis, Large Language Model, Prospect Theory},
}

\begin{document}

\twocolumn[
	\begin{@twocolumnfalse}
		\maketitle
\vspace{-3em}
\begin{abstract}
Large language models (LLMs) are now widely used across many fields, including marketing research. Sentiment analysis, in particular, helps firms understand consumer preferences. While most NLP studies classify sentiment from review text alone, marketing theories, such as prospect theory and expectation--disconfirmation theory, point out that customer evaluations are shaped not only by the actual experience but also by additional reference points. This study therefore investigates how the content and format of such supplementary information affect sentiment analysis using LLMs. 
We compare natural language (NL) and JSON-formatted prompts using a lightweight 3B parameter model suitable for practical marketing applications.
Experiments on two Yelp categories (Restaurant and Nightlife) show that the JSON prompt with additional information outperforms all baselines without fine-tuning: Macro-F1 rises by 1.6\% and 4\% while RMSE falls by 16\% and 9.1\%, respectively, making it deployable in resource-constrained edge devices.
Furthermore, a follow-up analysis confirms that performance gains stem from genuine contextual reasoning rather than label proxying. This work demonstrates that structured prompting can enable smaller models to achieve competitive performance, offering a practical alternative to large-scale model deployment.
\end{abstract}
\vspace{0.5em}
\vspace{2em}
	\end{@twocolumnfalse}
]

\section{Introduction}
\subsection{Background}
In recent years, with the rapid advancements in large language models \citep[LLMs][]{gpt3}, both industrial and academic areas in wide range of domains have utilized LLMs for data analytics, automation, and decision support. In particular, due to their high applicability in textual data, many studies have implemented sentiment analysis using LLMs  \citep{llm_sentiment_review}. 

LLMs indeed demonstrate remarkable capabilities in understanding textual context; however, the actual `context' is referred to as a relationship between the tokens, which is captured through Transformer \citep{transformer} and attention mechanisms \citep{attention} and most existing approaches limit their analysis to the linguistic context within review texts alone. Regarding real-world marketing applications, the actual context of consumer evaluation contains the factors which extend far beyond the written review, such as past purchasing patterns, prior experiences with the business, comparative evaluations against competitors, and opinions from social media. 

This gap is particularly relevant in customer relationship management \citep[CRM][]{loyalty,loyalty_measurement}, where understanding customer sentiment accurately drives business decisions. Marketing research has long established through prospect theory \citep{prospect_theory} and expectation-disconfirmation theory \citep[EDT][]{edt} that consumers evaluate experiences relative to these broader reference points. This insight remains largely unexplored in LLM-based sentiment analysis.

Furthermore, in practical deployment scenarios, particularly for real-time recommendation systems, two critical challenges emerge. First, computational efficiency is paramount—many businesses cannot deploy large models (70B+ parameters) due to latency and infrastructure constraints. Second, despite having rich contextual data (user histories, business metrics), current methods fail to effectively incorporate this information into LLM-based sentiment analysis.

\subsection{Research Gap}
Despite LLMs' ability to process diverse input formats, sentiment analysis studies predominantly focus on review text alone \citep{effect_of_rating1,effect_of_rating2,effect_of_rating3}. However, real-world platforms possess rich contextual information. From these gaps, we sequentially derive four research questions (RQs 1--4):
\begin{itemize}
\item[RQ1] Reference-point utilization: Does supplying user- and business-average ratings actually help an LLM classify sentiment more accurately?
\item[RQ2] Prompt format: If the same information is presented in a machine-readable structure or plain texts, does the prompt format affect the model performance?
\item[RQ3] Proxy effect: If supplying the reference points improve accuracy, is it due to their implicit encoding of the ground-truth labels?
\item[RQ4] Reference interactions: How do interactions between multiple reference points affect prediction accuracy?
\end{itemize}

We address these RQs by three experimental studies. In Study 1, we set up two approaches for the prompts: natural-language (NL) and machine-readability (JSON), and several combinations of contextual factors: user average (U), business average (B), and other attributes (O). We compare the model performance across these models. In Study 2, we test whether such information reflect the label; average ratings may act as a proxy of ground truth. Finally, in Study 3, we further examine how accuracy changes according to the interactions of those two reference points. We progressively explore not only whether reference points improve performance, but also how they function within the model's inference process.

The remainder of this study is constituted as follows: Section 2 reviews related studies, Section 3 outlines the model construction, and Section 4 presents empirical analyses. We discuss the key findings and implications in Section 5. Finally, we list our research limitations in Section 6.

\section{Related Study}
\subsection{Sentiment Analysis}
Sentiment analysis has been conducted with various methodologies, including lexicon-based models \citep{vader}, machine learning approaches with embeddings \citep{word2vec,fasttext}, and deep neural networks such as BERT and RoBERTa \citep{bert,roberta}.

Recently, LLM-based approaches have gained attention for sentiment analysis \citep{llm_sentiment_review}. Models like GPT \citep{gpt3} and Llama \citep{llama} demonstrate superior performance compared to fully-supervised models and even fine-tuned RoBERTa \citep{gpt35_sentiment,llm_sentiment_review}. Key advantages include broad applicability due to pre-training and ability to process raw text without extensive preprocessing, achieving high accuracy without fine-tuning.

However, existing approaches predominantly focus on review text alone, overlooking rich contextual information available in real-world platforms.

\subsection{A Role of Reference Point in Service Evaluation}
In the field of marketing, extensive research has examined how consumers evaluate the product and service quality. Notable frameworks include prospect theory and EDT. Prospect theory posits that consumers evaluate services by comparing their actual experience to a pre-established reference point \citep{prospect_theory}. If their experience falls short of this reference point, they tend to feel dissatisfied; conversely, if it exceeds the reference point, satisfaction is more likely. Furthermore, EDT explains the evaluation process from two perspectives. Absolute evaluation involves assessing whether the perceived quality meets a fixed standard, while relative evaluation is based on comparing prior expectations with the perceived quality \citep{edt}. In both cases, prior expectations play a significant role in determining overall customer satisfaction.

Prior studies on a wide range of products and services have examined the factors that affect or shape expectations. Some of the key factors include consumers’ past experiences \citep{edt, strategic_expectation, satisfaction_hotelduration, satisfaction_JOM}, reputations provided by other customers \citep{satisfaction_relative, satifaction_quickCasualRestaurant, satisfaction_restaurantEWOM}, and perceived value \citep{satifaction_quickCasualRestaurant, satisfaction_fastfoodPerceivedValue}. In particular, reputations—such as an average rating and the helpful reviews from other consumers—serve as important reference points when comparing prior expectations with actual experiences.

However, as noted above, because sentiment analysis is predominantly based on actual review texts, few existing studies have taken these supplementary factors into account. To more accurately capture customers’ preferences, it is crucial to incorporate such information into sentiment analysis. Therefore, we propose a framework that effectively leverages this additional information within LLMs.

\section{Proposed Model}
\subsection{Pre-trained Model}
To implement LLM-based sentiment analysis, we adopt Llama 3.2 with 3 billion (3B) parameters and is instruction-tuned (Llama-3.2-3B-Instruct\footnote{https://huggingface.co/meta-llama/Llama-3.2-3B-Instruct}). Llama family has been widely adopted in sentiment analysis \citep{niimi_nldb}. In particular, the 3B architecture is a lightweight model which is capable of on-device processing with maintaining privacy. For hyper-parameters, this study adopt temperature=1.0. The model will stop inference after generating one token (equivalent to underbar in the prompt).

\subsection{Prompting Strategy}
To obtain sentiment values, we use text-completion function which the model completes the continuous texts of the given prompt. 

\begin{figure}[h]
   \begin{center}
   \fbox{\includegraphics[width=0.95\linewidth]{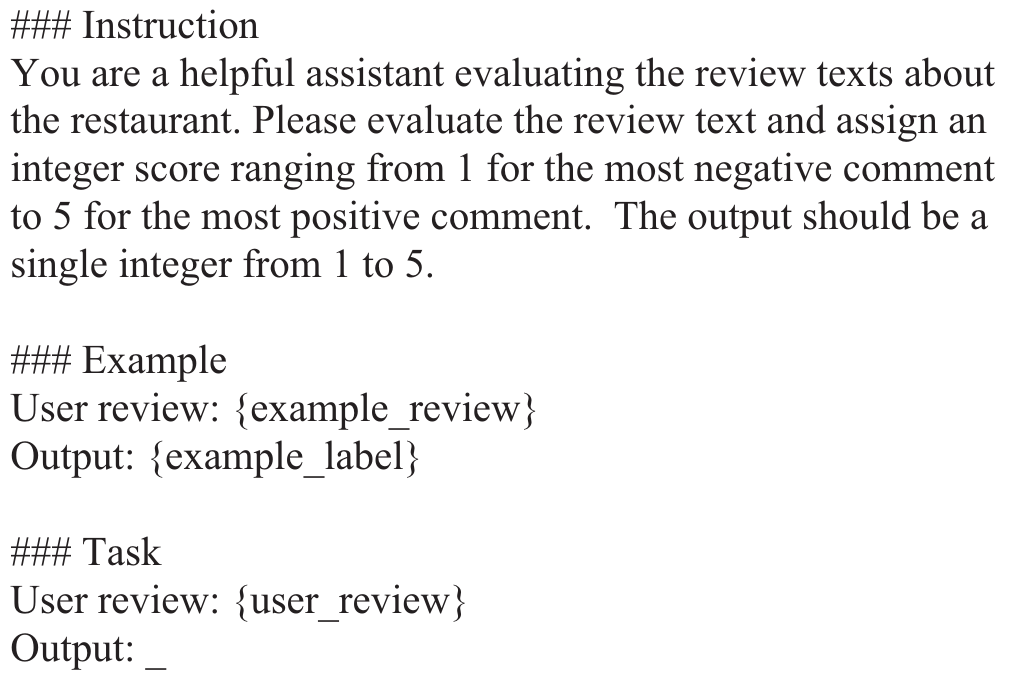}}
   \caption{Basic prompt for sentiment analysis (one-shot model)}\label{fig:instruction}
   \end{center}
\end{figure}

As shown in \citep{fewshots}, first, one-shot model is significantly outperforms the zero-shot model. Additionally, few-shot prompt clearly outperforms the one-shot; however, In this study, using multiple examples introduces additional variables (e.g., the reference points of example reviews, their alignment patterns, example selection biases), making it difficult to interpret core findings regarding structured contextual information. Therefore, we use one-shot prompt (Fig. \ref{fig:instruction}) to maintain experimental clarity.

\subsection{Displaying Supplementary Information}
To display additional information, the method of presentation needs to be discussed. Learning structural information is highly dependent on in-context learning, which means that locating the explanations before the tabular data improves the understanding of the structure \cite{tabularLLM}. Therefore, we set up and compare two display methods, text and JSON format. We include the supplementary information, such as the average user rating and the average venue rating after the review section of the prompt. 

\begin{figure}[h]
    \centering
    \fbox{\includegraphics[width=0.95\linewidth]{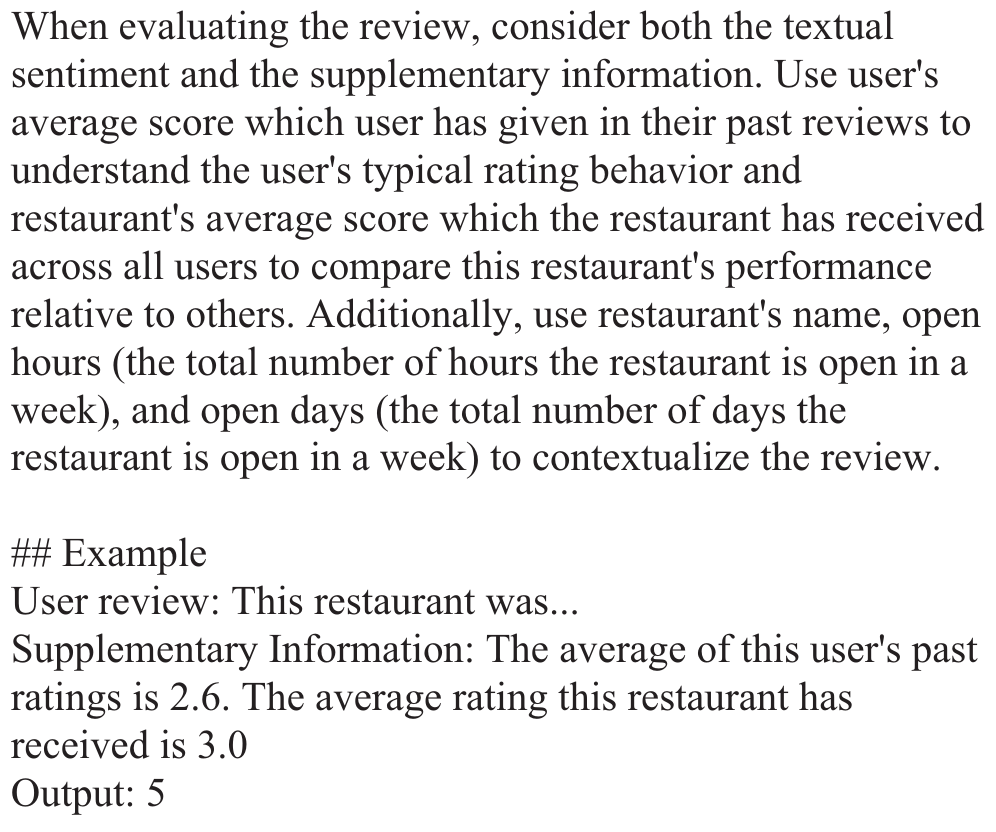}}
    \caption{Natural language (NL) prompt for supplementary information}\label{fig:text_style}
\end{figure}

\begin{figure}[h]
    \centering
    \fbox{\includegraphics[width=0.95\linewidth]{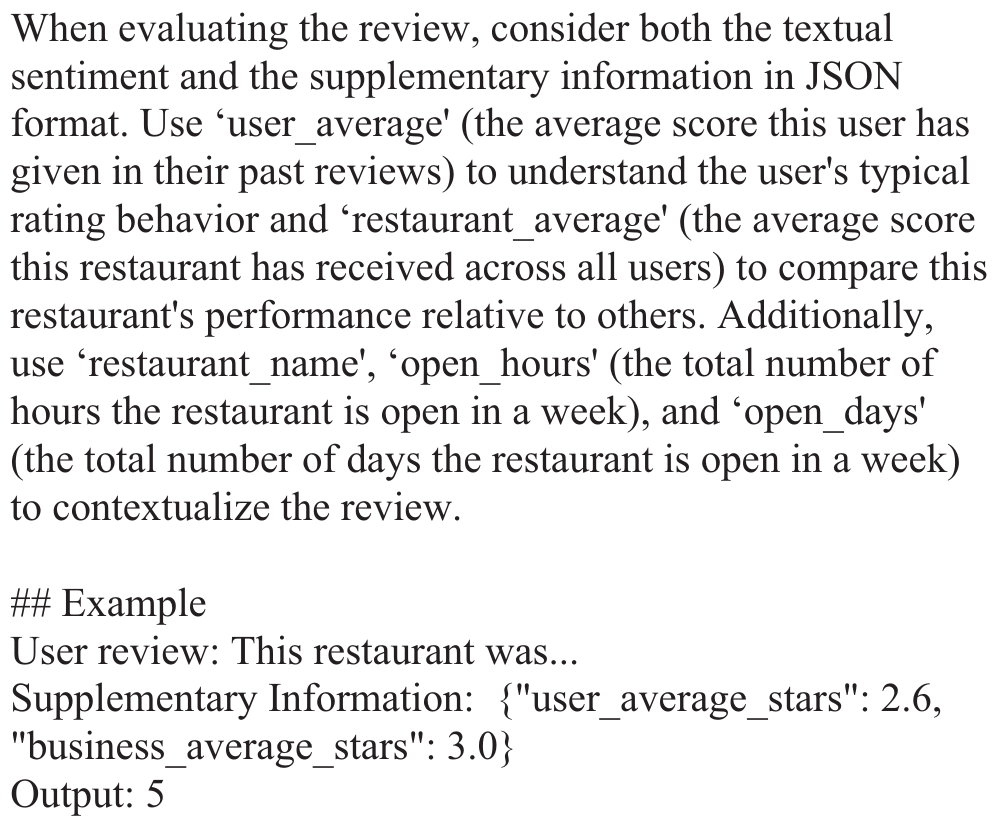}}
    \caption{Structured (JSON) prompt for supplementary information}\label{fig:json_style}
\end{figure}

\paragraph{Display Method (NL)}
First, we adopt text method, which displays the average evaluations with the explanations. With this method, although the prompt gets longer, any forms of information, including texts and numeric values, can be input into the model as long as the information can be explained in natural language. Thus, the necessary amount of computing resources becomes larger. Therefore, we set up the explanations and the input as Fig. \ref{fig:text_style}.

\paragraph{Display Method (JSON)} 
Second, we adopt JSON method, which displays supplementary information with JSON format. This method can provide information in the structured form. One study \cite{tabularLLM} which investigates the impact of using table data, such as CSV, JSON, and HTML, on the understanding of the information indicates that LLMs can comprehend the contents of the table unless the data structure is not too complex and the ability is improved when explanations are added before the structured input. Therefore, we set up the explanations and the structured input as Fig. \ref{fig:json_style}.

\subsection{Dataset}
We adopt Yelp Open Dataset \cite{yelp} which contains the evaluations and reviews for the variety of establishments. Some marketing studies in NLP field \cite{yelp_nldb2024,niimi_nldb} have utilized Yelp dataset since . 
For the comprehensive analysis, we set up the two different groups for the analysis: Restaurant and Nightlife, which are extracted using the category tags given for the establishments (Table \ref{tab:category}). For each group, we randomly select 500 English reviews from unique users to unique restaurants for a test set, which means that no duplicated users and establishments exist across the overall samples. 

\begin{table}[h]
\begin{center}
\caption{Selected and excluded category tags for each dataset. We avoid duplicate samples between datasets and select business with fixed addresses.}\label{tab:category}
\scalebox{0.9}{
\begin{tabular}{
wl{0.3cm}
wl{1.8cm}wl{4cm}
}
\toprule
&\ccol{1}{Selected} & \ccol{1}{Excluded}\\
\midrule
\multicolumn{2}{l}{\bf Restaurant}\\
&Restaurant & Fast Food, Food Truck,\\
&& Bar, Nightlife\\
\midrule
\multicolumn{2}{l}{\bf Nightlife}\\
&Bar,& Fast Food, Food Truck\\
& Nightlife  & \\
\bottomrule
\end{tabular}
}
\end{center}
\end{table}

For preprocessing the review texts, we at least remove the line break codes to maintain the format of the prompt. Table \ref{tab:statistics} shows the summary statistics of the dataset. A number of tokens is counted with tiktoken \cite{tiktoken} which is adopted in Llama 3. As shown in the statistics, some samples have significantly long texts.

\newcommand{\colwid}{1.0cm}
\begin{table}[h]
\begin{center}
\caption{Summary Statistics of the Datasets}\label{tab:statistics}
\scalebox{0.90}{
\begin{tabular}{
wl{0.5cm}
   wl{1.35cm}
   wr{\colwid}wr{\colwid}wr{\colwid}wr{\colwid}
   }
\toprule
&& \ccol{1}{Mean}&\ccol{1}{Std}&\ccol{1}{Min}&\ccol{1}{Max}\\
\midrule
\multicolumn{3}{l}{\bf Restaurant}\\
&Stars & 3.724 & 1.515 & 1~~~~~ & 5~~~~~ \\
&Chars &431.726 & 368.464 & 42~~~~~ & 2552~~~~~  \\
&Tokens & 98.744 & 85.544 & 9~~~~~ & 606~~~~~ \\
\midrule
\multicolumn{3}{l}{\bf Nightlife}\\
&Stars & 3.544 & 1.605 & 1~~~~~ & 5~~~~~  \\
&Chars & 511.082 & 484.695 & 65~~~~~ & 4998~~~~~ \\
&Tokens &117.104 & 112.465 & 15~~~~~ & 1118~~~~~  \\
\bottomrule
\end{tabular}
}
\end{center}
\end{table}

\section{Analyses and Results}
\newcommand{\modelwidth}{2.4cm}
\newcommand{\metricswidth}{0.9cm}
\newcommand{\ignore}[1]{}
\subsection{Study 1: Impact of Reference Points and Display Methods}
First, to address RQ1 (reference-point utilization) and RQ2 (prompt format), we implement sentiment analysis in restaurant and bar evaluations. We compare evaluation metrics across different models and assess the effectiveness of incorporating additional information in two categories. 
The supplementary information consists of following three elements. U: the user’s average rating, indicating the mean rating of the past evaluations given by the user on Yelp, B: the business’ average rating, indicating the mean rating which the restaurant has received from all users, and O: other contextual factors, indicating additional information both of textual and numerical attributes, such as the restaurant name, operating hours and the number of days the restaurant is open per week. Both U and B are expected to serve as reference points that affect user’s prior expectations. 

The LLM-based approach is evaluated with multiple variations, considering the type of supplementary information used and its machine-readability. Accordingly, the LLM-based models are categorized as follows: JSON-UBO / NL-UBO: Utilizing all supplementary information, presented in JSON format or natural language, respectively; JSON-UB / NL-UB: Incorporating only the average ratings; JSON-O / NL-O: Incorporating only contextual factors; and LLM (None): A baseline model without any supplementary information. We also establish four well-established baselines: BERT  \cite{bert}, DistilBERT \cite{distilbert}, RoBERTa \cite{roberta}, and DeBERTa \cite{deberta}. These pretrained models are fine-tuned for 5 epochs with additionally extracted 1000 training samples and the test performances are computed when the validation losses become the lowest. Since the proposed model employs 3B model which focused on lightweight and fast inferences, reference models are also base-sized (e.g., RoBERTa-Base-Uncased).

Sentiment analysis has not only the classification aspect but also the regression due to that the sentiment label is on the ordinal scale, which means that the magnitude of prediction error is important in addition to the concordance. Therefore, we adopt both Macro-F1 score and root mean square error (RMSE) for the evaluation metrics. For the baseline models we extracted an additional training set and fine-tuned each model. To prevent any data leakage, we ensured that both user IDs and store IDs are mutually exclusive between the training and test sets of the entire Yelp dataset. Under this constraint, we use at most 500 unique user--business pairs for our evaluation set.

\begin{table}[htb]
\begin{center}
\caption{Study 1 Results (Dataset 1: Restaurant). $\dagger$ indicates the statistically significant difference ($p<.05$) by two-sided McNemar test against LLM (None). Bold number indicates that the model surpasses all the reference models; shaded cells indicate the overall best value.}\label{tab:res1-1}
\begin{threeparttable}
\begin{tabular}{
wc{0.5cm}wl{\modelwidth}
wc{\metricswidth}wc{\metricswidth}wc{\metricswidth}
}
\toprule
 \multicolumn{2}{l}{$n=500$}& \ccol{1}{UBO} & \ccol{1}{UB} & \ignore{\ccol{1}{U} & \ccol{1}{B} &} \ccol{1}{O} \\
\midrule
\multirow{7}{*}{\rotatebox[origin=c]{90}{Macro-F1}}
&LLM (JSON) & \graycell\bf0.612$^\dagger$ &  \bf0.598 & \ignore{0.551 &  \bf0.621 &}  \bf0.588 \\
&LLM (NL) & \bf0.593$^\dagger$ &  \bf0.599 & \ignore{0.553 &  \bf0.641 &} 0.524 \\
\cmidrule(lr){2-5}
&LLM (None) & 0.587 \\
&DeBERTa & 0.538 \\
&RoBERTa & 0.533 \\
&BERT & 0.474 \\
&DistilBERT & 0.465 \\
\midrule
\multirow{7}{*}{\rotatebox[origin=c]{90}{RMSE}}
&LLM (JSON) & \graycell \bf0.564 &  \bf0.616 & \ignore{\bf0.660 &  \bf0.642 &}  \bf0.647\\
&LLM (NL) &  \bf0.620 &  \bf0.624 & \ignore{\bf0.612 &  \bf0.628 &}  0.686 \\
\cmidrule(lr){2-5}
&LLM (None) & 0.675 \\
&DeBERTa & 0.703 \\
&RoBERTa & 0.742 \\
&BERT & 0.758 \\
&DistilBERT & 0.804 \\
\bottomrule
\end{tabular}
\begin{tablenotes}[para,flushleft,online,normal] %(default:normal)
\end{tablenotes}
\end{threeparttable}
\end{center}
\end{table}

\begin{table}[htb]
\begin{center}
\caption{Study 1 Results (Dataset 2: Nightlife). $\dagger$ indicates the statistically significant difference ($p<.05$) by two-sided McNemar test against LLM (None). Bold number indicates that the model surpasses all the reference models; shaded cells indicate the overall best value.}\label{tab:res1-2}
\begin{threeparttable}
\begin{tabular}{
wc{0.5cm}wl{\modelwidth}
wc{\metricswidth}wc{\metricswidth}wc{\metricswidth}
wc{\metricswidth}wc{\metricswidth}
}
\toprule
 \multicolumn{2}{l}{$n=500$}& \ccol{1}{UBO} & \ccol{1}{UB} & \ignore{\ccol{1}{U} & \ccol{1}{B} &} \ccol{1}{O} \\
\midrule
\multirow{7}{*}{\rotatebox[origin=c]{90}{Macro-F1}}
&LLM (JSON) &\graycell\bf0.635$^\dagger$ & 0.622 &  0.592 \\
&LLM (NL) & 0.602 &  \bf0.628 & 0.580 \\
\cmidrule(lr){2-5}
&LLM (None) & 0.526 \\
&DeBERTa & 0.523 \\
&RoBERTa & 0.625 \\
&BERT & 0.574 \\
&DistilBERT & 0.481 \\
\midrule
\multirow{7}{*}{\rotatebox[origin=c]{90}{RMSE}}
&LLM (JSON) & \graycell\bf0.597 & \bf0.613 & 0.665 \\
&LLM (NL) & 0.672 & \bf0.647  & 0.666 \\
\cmidrule(lr){2-5}
&LLM (None) & 0.709 \\
&DeBERTa & 0.688 \\
&RoBERTa & 0.657 \\
&BERT & 0.668 \\
&DistilBERT & 0.746 \\
\bottomrule
\end{tabular}
\begin{tablenotes}[para,flushleft,online,normal] %(default:normal)
\end{tablenotes}
\end{threeparttable}
\end{center}
\end{table}

Table \ref{tab:res1-1} and \ref{tab:res1-2} report the results for the Restaurant and Nightlife datasets, respectively. 
First, among the two datasets, JSON-UBO achieves the highest score in both datasets and improves significantly over LLM (None), which receives no supplementary information. Among the reference models, the strongest baseline differs by domain: LLM (None) ranks first in Restaurant, whereas RoBERTa-Base leads in Nightlife. In general, RoBERTa and DeBERTa outperform BERT, followed by DistilBERT. 

Comparing the models within each display format, first, JSON prompts consistently improve performance as more information is added. These results indicate that, supplying a reference point in a machine-friendly format helps the model capture the complex relationships among factors, enabling effective inference. Increasing information from JSON-UB to JSON-UBO leads particularly to reduce RMSE in both datasets. As a result, prediction accuracy rises; for Restaurant, Macro-F1 rises by 4.3\% (from 0.587 to 0.612) and RMSE reduces by 16.44\% (from 0.675 to 0.564) relative to both LLM (None). For Nightlife, the improvements are even larger relative to LLM (None), +20.7\% (from 0.526 to 0.635) and -15.8\% (from 0.709 to 0.597), respectively, and still effective compared with RoBERTa (+1.6\% / -9.1\%). This result aligns with prospect theory and expectation-disconfirmation theory, providing the clear answer to RQ1. The user- and business-level average ratings act as reference points for the LLM and improve the prediction accuracy.

By contrast, the results of NL prompts do not follow this pattern; increasing information from NL-UB to NL-UBO does not contribute on the performance, and in Nightlife the accuracy even decreases below the best baseline. This suggests that, although LLMs can process natural language using the large context window, the models still struggle to capture their complex relationship particularly when large quantities of contextual factors are embedded as plain text. 

These empirical results also provide the response to RQ2. Supplying additional information in a machine-readable structure allows the LLMs to effectively utilize those reference points and contextual factors for the prediction. 

\subsection{Study 2: Relationship with the Expectation}
From the results of Study 1, a remaining concern is that these reference points may have worked as proxies for the labels. Therefore, to address RQ3 (proxy effect), Study 2 investigates whether model performance decreases as the review score diverges from these reference points according to the extent of the gaps. We use the JSON-UBO results for the examination.

Since prior expectations are formed based on the user's past behavior and the reputations, we treat such average score as indicators of prior expectations. We define expectation--evaluation gap for both user (U) and business (B) average from user $i$ to store $j$ as follows:
\begin{align}
  gap^{(U)}_{i,j} &= rating_{i,j} - user\_average_i\\
  gap^{(B)}_{i,j} &= rating_{i,j} - business\_average_j
\end{align}
The data set is divided into five bins according to the extent of gaps, yielding groups that range from “far below expectations” to “far above expectations.” For each bin, we measure the Micro-F1 and RMSE. If the averages were merely proxies for the labels, performance would peak in the middle bin (where the gap is smallest) and decline sharply as the gap widens. 

The results are shown in Table \ref{tab:res2-1} (Restaurant) and Table \ref{tab:res2-2} (Nightlife). The leftmost group includes cases where the actual rating falls below expectations, while the rightmost group consists of cases where the actual rating exceeds expectations. 
\begin{table}[htb]
\begin{center}
\caption{Study 2 Results (Dataset 1: Restaurant). Bold number indicates that the group surpasses the expectation-met group. Shaded cells indicate the overall best value.}\label{tab:res2-1}
\begin{threeparttable}
\begin{tabular}{
wl{1.5cm}wr{0.7cm}wr{0.7cm}wr{0.7cm}wr{0.7cm}wr{0.7cm}
}
\toprule
& \ccol{5}{Expectation}\\
 \cmidrule{2-6}
& \ccol{2}{below~~~$\leftarrow$} & \ccol{1}{met} & \ccol{2}{$\rightarrow$~~~beyond} \\
\midrule
\multicolumn{4}{l}{\bf User Average}\\
 $gap^{(U)}_{i,j}$ & -1.788 & -0.185 & 0.224 & 0.748 & 1.653 \\
 Micro-F1 & \bf0.690 & 0.636 & 0.667 & \graycell\bf0.890 & \bf0.870 \\
 RMSE & 0.700 & 0.621 & 0.619 & \graycell\bf0.374 & \bf0.436 \\
\midrule
\multicolumn{4}{l}{\bf Business Average}\\
 $gap^{(B)}_{i,j}$ & -2.130 & -0.640 & 0.415 & 0.945 & 1.565 \\
 Micro-F1 & 0.760 & 0.450 & 0.760 & \bf0.860 & \graycell\bf0.910 \\
 RMSE & 0.624 & 0.819 & 0.490 & \bf0.447 & \graycell\bf0.300 \\
\bottomrule
\end{tabular}
\begin{tablenotes}[para,flushleft,online,normal] %(default:normal)
\end{tablenotes}
\end{threeparttable}
\end{center}
\end{table}

First, in the Restaurant category, prediction performances increase in the upper two quantile groups for both the user's and store's average, compared to the middle group where the actual rating was close to the reference point. This suggests that the most accurate predictions were made when the actual experience exceeded prior expectations. In particular, when the experience was beyond the expectation, the performance improved by +25.1\% for Micro-F1 and -39.6\% for RMSE in user-average compared with the middle group while +13.2\% for Micro-F1 and -8.8\% for RMSE in business-average.

\begin{table}[htb]
\begin{center}
\begin{threeparttable}
\caption{Study 2 Results (Dataset 2: Nightlife). Bold number indicates that the group surpasses the expectation-met group. Shaded cells indicate the overall best value.}\label{tab:res2-2}
\begin{tabular}{
wl{1.5cm}wr{0.7cm}wr{0.7cm}wr{0.7cm}wr{0.7cm}wr{0.7cm}
}
\toprule
& \ccol{5}{Expectation}\\
 \cmidrule{2-6}
& \ccol{2}{below~~$\leftarrow$} & \ccol{1}{met} & \ccol{2}{$\rightarrow$~~beyond} \\
\midrule
\multicolumn{4}{l}{\bf User Average}\\
 $gap^{(U)}_{i,j}$ & -2.083 & -0.312 & 0.133 & 0.523 & 1.375 \\
 Micro-F1 & \bf0.743 & \bf0.709 & 0.700 & \bf0.752 & \graycell\bf0.798 \\
 RMSE & \graycell\bf0.507 & 0.660 & 0.548 & 0.554 & 0.611 \\
\midrule
\multicolumn{4}{l}{\bf Business Average}\\
 $gap^{(B)}_{i,j}$ & -2.465 & -0.995 & 0.350 & 0.975 & 1.520 \\
 Micro-F1 & \graycell\bf0.830 & 0.520 & 0.710 & \bf0.850 & \bf0.800 \\
 RMSE & \graycell\bf0.412 & 0.911 & 0.566 & \bf0.510 & \bf0.447 \\
\bottomrule
\end{tabular}
\begin{tablenotes}[para,flushleft,online,normal] %(default:normal)
\end{tablenotes}
\end{threeparttable}
\end{center}
\end{table}

Next, in the Nightlife category, as in Study 1, merely meeting expectations did not consistently lead to better predictions. However, unlike the Restaurant category, prediction accuracy improved not only when experiences exceeded expectations, but also when they fell significantly short. Notably, the highest accuracy was observed in the group where the actual rating was far below the prior expectation (the leftmost group).

Although there are substantial differences in business nature and customer behavior between restaurants and nightlife venues, at least, we do not confirm that the performance metrics increase in the group with the closest reference points to actual labels. In both cases, reference points do not simply function as proxies for the correct labels, but rather as relative evaluation values in inference, meaning that they function literally as reference points, which is a strong answer to RQ3.

\subsection{Study 3: Error Analysis}
To further understand how the model interacts with the reference points, we finally combine user and business average scores to create a 5×5 matrix where each cell represents the Micro-F1 score.

\begin{table}[h]
\begin{center}
\caption{Error analysis by user average (UA) and business average (BA) for restaurant dataset}\label{tab:error1}
\scalebox{0.98}{
\begin{tabular}{
wl{0.35cm}
lccccc
}
\toprule
\multicolumn{3}{l}{Restaurant}&\ccol{3}{BA}&\\
\cmidrule(lr){3-7}
& & 1 & 2 & 3 & 4 & 5  \\
\midrule
&1 & - & 1.000 & 0.812 & 0.750 & - \\
&2 & 1.000 & 0.833 & 0.771 & 0.643 & 0.000 \\
UA&3 & 1.000 & 0.643 & 0.688 & 0.716 & 0.000 \\
&4 & - & 0.250 & 0.679 & 0.816 & 0.750 \\
&5 & - & 1.000 & 0.909 & 1.000 & 1.000 \\
\bottomrule
\end{tabular}
}
\end{center}
\end{table}

\begin{table}[h]
\begin{center}
\caption{Error analysis by user average (UA) and business average (BA) for nightlife dataset}\label{tab:error2}
\scalebox{0.98}{
\begin{tabular}{
wl{0.35cm}
lccccc
}
\toprule
\multicolumn{3}{l}{Nightlife}&\ccol{3}{BA}&\\
\cmidrule(lr){3-7}
& & 1 & 2 & 3 & 4 & 5  \\
\midrule
&1 & - & 1.000 & 0.920 & 1.000 & - \\
&2 & - & 0.714 & 0.680 & 0.545 & - \\
UA &3 & - & 0.750 & 0.623 & 0.756 & - \\
&4 & - & 0.556 & 0.682 & 0.746 & 1.000 \\
&5 & - & 1.000 & 1.000 & 1.000 & 1.000 \\
\bottomrule
\end{tabular}
}
\end{center}
\end{table}

Table \ref{tab:error1} (Restaurant) and \ref{tab:error2} (Nightlife) show the results. First, in both categories, the model achieves highest performance (100\% in most cases) when UA is 5. Second, two different reference points show a clear interaction for the prediction. The accuracy tends to improve when two reference points align (along the diagonal), indicating that, when user's past evaluation is close to  other consumers' average ratings, the actual rating becomes easier to predict. 
Notably, in some combinations, the accuracy results in 0\%. This indicates cases where conflicting references make prediction challenging. However, as shown in Study 1, this accuracy even outperforms other models. Therefore, our approach can identify unreliable or difficult samples to predict based on reference point conflicts. These results clearly answer RQ4.

These interaction patterns enable practical deployment strategies. Companies can employ adaptive inference where samples with aligned reference points (UA$\approx$BA) are processed on-device environment, while conflicting cases are routed to larger cloud-based models. Additionally, low-confidence predictions can be systematically collected as training data for fine-tuning domain-specific models, enabling continuous performance improvement. 

\section{Conclusion}
\subsection{Key Findings}
In this study, we enhance LLM-based sentiment analysis by incorporating supplementary information as reference points and other contextual factors, based on prospect theory and EDT. 

Study 1 compared two prompting strategies (NL and JSON) with multiple combinations of contextual information. The JSON-UBO model significantly outperformed both NL prompts and four strong baselines. Notably, while JSON prompts showed consistent gains with increasing information, NL prompts failed to leverage the same contexts despite the same context window.

Study 2 addressed the potential concern about reference points serving as label proxies. Accuracy improved more for reviews whose ratings deviated from the average than for those close to the average, indicating that the model was not simply copying the reference points for JSON-UBO model.

Study 3 revealed that the interactions of two different reference points affect the model performance. Accuracy improved when those ratings align while conflicting references indicate inherently challenging cases.

These findings comprehensively answer our research questions. RQ1 (Effect of reference points): We demonstrated that the proposed model with U/B/O information significantly improves performance with 4.3--20.7\% gains in Macro-F1 and 15.8--16.4\% reductions in RMSE over baselines without fine-tuning. RQ2 (Effect of machine readability): NL prompts fail to leverage complex information, highlighting the importance of prompt design even for models with large context windows. RQ3 (Validation of contextual reasoning): Follow-up analysis confirms that model performance improves when ratings deviate from expectations, indicating that reference points assist contextual inference rather than serving as mere label proxies. RQ4 (Reference nteractions): We revealed that aligned reference points improve prediction accuracy, while conflicting reference points indicate inherently challenging prediction cases.

\subsection{Implications}
This study has both academic and practical implications.

\paragraph{Academic.}
First, by incorporating the theoretical approach into the LLM-based sentiment analysis, model performance significantly improved, indicating that LLM's rich capability to handle complex context contributes to the predictions. By using JSON format, it is possible to input various information into LLMs, and combining more abundant information may further improve prediction accuracy.
Second, even if the amount of information is same across the several prompts, the results of the inferences, including the prediction and performance, vary depending on the display methods, despite the large context window of modern LLMs. Second, simple scalar values, such as 1--5 star averages, can be used directly in the JSON prompt; no discretization or embedding tricks are required. These suggest that we can flexibly employ various factors, including textual and numeric information, into sentiment analysis.
Furthermore, although we employ sentiment analysis for the model verification, the proposed approach using JSON-based contextual information is transferable in wide domain of document classification task, including marketing analysis.

\paragraph{Practical.}
Since the proposed method only relies on prompt construction, companies can feed existing database contents to LLMs with JSON prompt and immediately construct their own extended models. 
As shown in the results, our approach achieves RMSE of 0.564 (restaurant) and 0.597 (nightlife), meaning the average prediction error is less than 1-star on a 5-point scale. This level of accuracy is sufficient for practical applications such as the simple recommendation systems, where distinguishing between adjacent rating categories (e.g., 4 vs 5 stars) is often less critical than identifying overall sentiment polarity. Achieving this performance with a 3B parameter scale without fine-tuning indicates that the company can immediately deploy the recommendation agent on edge application environments combined with the rich customer database. Furthermore, our error analysis revealed that samples with aligned reference points can be accurately predicted while cases with conflicting references could be routed to larger models or LLM-based ensemble strategy \citep{llm_ensemble2,llm_ensemble_finance,niimi_nldb}. This enables the energy efficient processing where computational resources are allocated based on prediction difficulty.

\subsection{Limitations}
This study has several limitations. First, while our approach is grounded in prospect theory and EDT, we did not empirically test whether the psychological mechanisms underlying these theories actually explain the model's improved performance. 

Second, our experiments were conducted using only a single model architecture (Llama-3.2-3B-Instruct). The effectiveness of structured prompting may vary significantly across different model families and sizes, limiting the generalizability of our findings. Particularly, we cannot conclude whether the benefits of JSON formatting extend to larger models or different architectures.

Third, our evaluation is restricted to English-language reviews from two categories within the Yelp Open Dataset \citep{yelp}. The effectiveness on other languages, domains, or review platforms with different characteristics of supplementary information remains untested. 

Finally, while we argue that our approach is computationally efficient due to the absence of fine-tuning, we did not provide quantitative measurements of inference time or memory usage across different prompt formats for the actual inferences.

\section*{Acknowledgments}
The dataset used in this study consisted of open data for academic use. As no additional information was collected, the author did not obtain any information that could lead to the identification of individuals. The large language models used to construct the analysis model were licensed for commercial and academic use. Both the dataset and models were managed and used in an appropriate environments that comply with the terms of use of the companies from which it was made available.

This study is supported by JSPS KAKENHI (Grant Number: 24K16472).

\bibliographystyle{unsrt}
\bibliography{supplLLM}

\end{document}